\documentclass[11pt,a4paper]{article}
\usepackage[hyperref]{acl2019}

\usepackage{times}
\usepackage{latexsym}
\usepackage{multirow}
\usepackage{array}

\usepackage{graphicx}
\usepackage{booktabs}
\usepackage{numprint}
\usepackage{amssymb}
\usepackage{enumerate}

\usepackage{xspace}
\usepackage{textcomp}

\usepackage{url}
\usepackage{float}
\usepackage{setspace}
\restylefloat{table}

\aclfinalcopy

\title{Naver Labs Europe's Systems for the WMT19\\ Machine Translation Robustness Task}

\author{
	Alexandre B\'erard \qquad
	Ioan Calapodescu \qquad
        Claude Roux
	\\[.4em] Naver Labs Europe\\first.last@naverlabs.com
}

\date{}
\newcolumntype{H}{>{\setbox0=\hbox\bgroup}c<{\egroup}@{}}

\begin{document}

\maketitle

\begin{abstract}
This paper describes the systems that we submitted to the WMT19 Machine Translation robustness task. This task aims to improve MT's robustness to noise found on social media, like informal language, spelling mistakes and other orthographic variations. The organizers provide parallel data extracted from a social media website\footnote{\url{https://www.reddit.com}} in two language pairs: French-English and Japanese-English (in both translation directions). The goal is to obtain the best scores on unseen test sets from the same source, according to automatic metrics (BLEU) and human evaluation.
We proposed one single and one ensemble system for each translation direction. Our ensemble models ranked first in all language pairs, according to BLEU evaluation.
We discuss the pre-processing choices that we made, and present our solutions for robustness to noise and domain adaptation.
\end{abstract}

\section{Introduction}

Neural Machine Translation (NMT) has achieved impressive results in recent years, especially on high-resource language pairs \cite{vaswani_2017,edunov_2018}, and has even lead to some claims of human parity \cite{hassan_2018}.\footnote{These claims were discussed at WMT by \citet{toral_2018}.}

However, \citet{belinkov_2018} show that NMT is brittle, and very sensitive to simple character-level perturbations like letter swaps or keyboard typos. They show that one can make an MT system more robust to these types of synthetic noise, by introducing similar noise on the source side of the training corpus. \citet{sperber_2017} do similar data augmentation, but at the word level and so as to make an MT model more robust to Automatic Speech Recognition errors (within a speech translation pipeline). \citet{cheng_2018} propose an adversarial training approach to make an encoder invariant to word-level noise. \citet{karpukhin_2019} propose to inject aggressive synthetic noise on the source side of training corpora (with random char-level operations: deletion, insertion, substitution and swap), and show that this is helpful to deal with natural errors found in Wikipedia edit logs, in several language pairs.

\citet{michel_2018} release MTNT, a real-world noisy corpus, to help researchers develop MT systems that are robust to natural noise found on social media. The same authors co-organized this task \cite{WMT_robustness_2019}, in which MTNT is the primary resource. \citet{vaibhav_2019} show that back-translation (with a model trained on MTNT) and synthetic noise (that emulates errors found in MTNT) are useful to make NMT models more robust to MTNT noise.

This task aims at improving MT's robustness to noise found on social media, like informal language, spelling mistakes and other orthographic variations. We present the task in more detail in Section~\ref{section:task_description}. Then, we describe our baseline models and pre-processing in Section~\ref{section:baseline_models}. We extend these baseline models with robustness and domain adaptation techniques that are presented in Section~\ref{section:robustness_techniques}. Finally, in Section~\ref{section:results}, we present and discuss the results of our systems on this task.

\section{Task description}
\label{section:task_description}

The goal of the task is to make NMT systems that are robust to noisy text found on Reddit, a social media, in two language pairs (French-English and Japanese-English) and both translation directions.
The evaluation will be performed on a blind test set (obtained from the same source), using automatic metrics and human evaluation. We present our final BLEU scores in Section~\ref{section:results}, while the human evaluation results are given in the shared task overview paper \cite{WMT_robustness_2019}.

\paragraph{MTNT}
\citet{michel_2018} crawled monolingual data from Reddit in three languages: English, French and Japanese, which they filtered to keep only the ``noisiest'' comments (containing unknown words or with low LM scores).

Then, they tasked professional translators to translate part of the English data to French, and part of it to Japanese. The Japanese and French data was translated to English. The resulting parallel corpora were split into train, valid and test sets (see Table~\ref{table:MTNT_data}). The test sets were manually filtered so as to keep only good quality translations. The data that was not translated is made available as monolingual corpora (see Table~\ref{table:monolingual_data}).

\paragraph{Other data}
In addition to the provided in-domain training and evaluation data, we are allowed to use larger parallel and monolingual corpora (see Tables~\ref{table:out_of_domain_parallel_data} and~\ref{table:monolingual_data}). For FR$\leftrightarrow$EN, any parallel or monolingual data from the WMT15 news translation task\footnote{\url{http://www.statmt.org/wmt15/translation-task.html}} is authorized. For JA$\leftrightarrow$EN, we are allowed the same data that was used by \citet{michel_2018}: KFTT, TED and JESC.

\paragraph{Challenges}

\citet{michel_2018} identified a number of challenges for Machine Translation of MTNT data, which warrant the study of MT robustness. Here is an abbreviated version of their taxonomy:

\begin{itemize}
	\setlength\itemsep{0em}
	\item Spelling and grammar mistakes: e.g., their/they're, have/of.
	\item Spoken language and internet slang: e.g., lol, \emph{mdr}, lmao, etc.
	\item Named entities: many Reddit posts link to recent news articles and evoke celebrities or politicians. There are also many references to movies, TV shows and video games.
	\item Code switching: for instance, Japanese text on Reddit contains many English words.
	\item Reddit jargon: words like ``downvote'', ``upvote'' and ``cross-post'',\footnote{The French-speaking Reddit community sometimes uses funny literal translations of these: ``bas-vote'', ``haut-vote'' and ``croix-poteau''.} and many acronyms like TIL (Today I Learned), OP (Original Poster), etc.
	\item Reddit markdown: characters like ``$\sim$'', ``*'' and ``\^{}" are extensively used for formatting.
	\item Emojis and emoticons  (``;-)'').
	\item Inconsistent capitalization: missing capital letters on proper names, capitalization for emphasis or ``shouting'', etc.
	\item Inconsistent punctuation.
\end{itemize}

\paragraph{Evaluation}
Automatic evaluation is performed with cased BLEU \cite{papineni_2002}, using SacreBLEU \cite{post_2018}.\footnote{\texttt{BLEU+case.mixed+numrefs.1+smooth.exp\\+tok.13a+version.1.3.1}} For English and French, the latter takes as input detokenized MT outputs and untokenized reference data. For Japanese, MT outputs and reference are first tokenized with Kytea\footnote{\texttt{kytea -model share/kytea/model.bin -out tok} (v0.4.7)} \cite{neubig_2011} before being processed by SacreBLEU (because it does not know how to tokenize Japanese).
The organizers will also collect subjective judgments from human annotators, and rank participants accordingly.

\begin{table}[t]
	\centering
	\begin{tabular}{lccc}
		\multirow{2}{*}{Lang pair} & \multirow{2}{*}{Lines} & \multicolumn{2}{c}{Words} \\
		& & Source & Target \\
		\hline
		JA$\to$EN & 6 506 & 160k & 155k \\
		EN$\to$JA & 5 775 & 339k & 493k \\
		FR$\to$EN & 19 161 & 794k & 763k \\
		EN$\to$FR & 36 058 & 1 014k & 1 152k \\
	\end{tabular}
	\vspace{-.2cm}
	\caption{Size of the MTNT training corpora. Word counts by Moses (fr/en) and Kytea (ja) tokenizers.}
	\label{table:MTNT_data}
\end{table}

\begin{table}[t]
	\centering
	\begin{tabular}{lccc}
		\multirow{2}{*}{Lang pair} & \multirow{2}{*}{Lines} & \multicolumn{2}{c}{Words} \\
		& & Source & Target \\
		\hline
		JA$\leftrightarrow$EN & 3.90M & 48.42M & 42.63M \\
		FR$\leftrightarrow$EN & 40.86M & 1 392M & 1 172M\\
	\end{tabular}
	\vspace{-.2cm}
	\caption{Size of the authorized out-of-domain parallel corpora in constrained submissions.}
	\label{table:out_of_domain_parallel_data}
\end{table}

\begin{table}[H]
	\centering
	\begin{tabular}{l|cc}
		Language & Corpus & Lines \\
		\hline
		Japanese & MTNT & 32 042 \\
		\hline
		\multirow{3}{*}{French} & MTNT & 26 485 \\
		& news-discuss & 3.84M \\
		& news-crawl & 42.1M \\
		\hline		
		\multirow{3}{*}{English} & MTNT & 81 631 \\
		& news-discuss & 57.8M \\
		& news-crawl & 118.3M \\
	\end{tabular}
	\vspace{-.2cm}
	\caption{Authorized monolingual data.}
	\label{table:monolingual_data}
\end{table}

\section{Baseline models}
\label{section:baseline_models}

This section describes the pre-processing and hyper parameters of our baseline models. We will then detail the techniques that we applied for robustness and domain adaptation.

\subsection{Pre-processing}

\paragraph{CommonCrawl filtering}

We first spent efforts on filtering and cleaning the WMT data (in particular CommonCrawl).

We observed two types of catastrophic failures when training FR$\to$EN models: source  sentence copy, and total hallucinations.

The first type of error (copy) is due to having sentence pairs in the training data whose reference ``translation'' is a copy of the source sentence. \citet{khayrallah_2018} show that even a small amount of this type of noise can have catastrophic effects on BLEU. We solve this problem by using a language identifier (\texttt{langid.py}, \citealp{lui_2012}) to remove any sentence pair whose source or target language is not right.

Then, we observed that most of the hallucinations produced by our models were variants of the same phrases (see Table~\ref{table:hallucination} for an example). We looked for the origin of these phrases in the training data, and found that they all come from CommonCrawl \citep{smith_2013}.

\begin{figure}
	\centering
	\includegraphics[width=8cm]{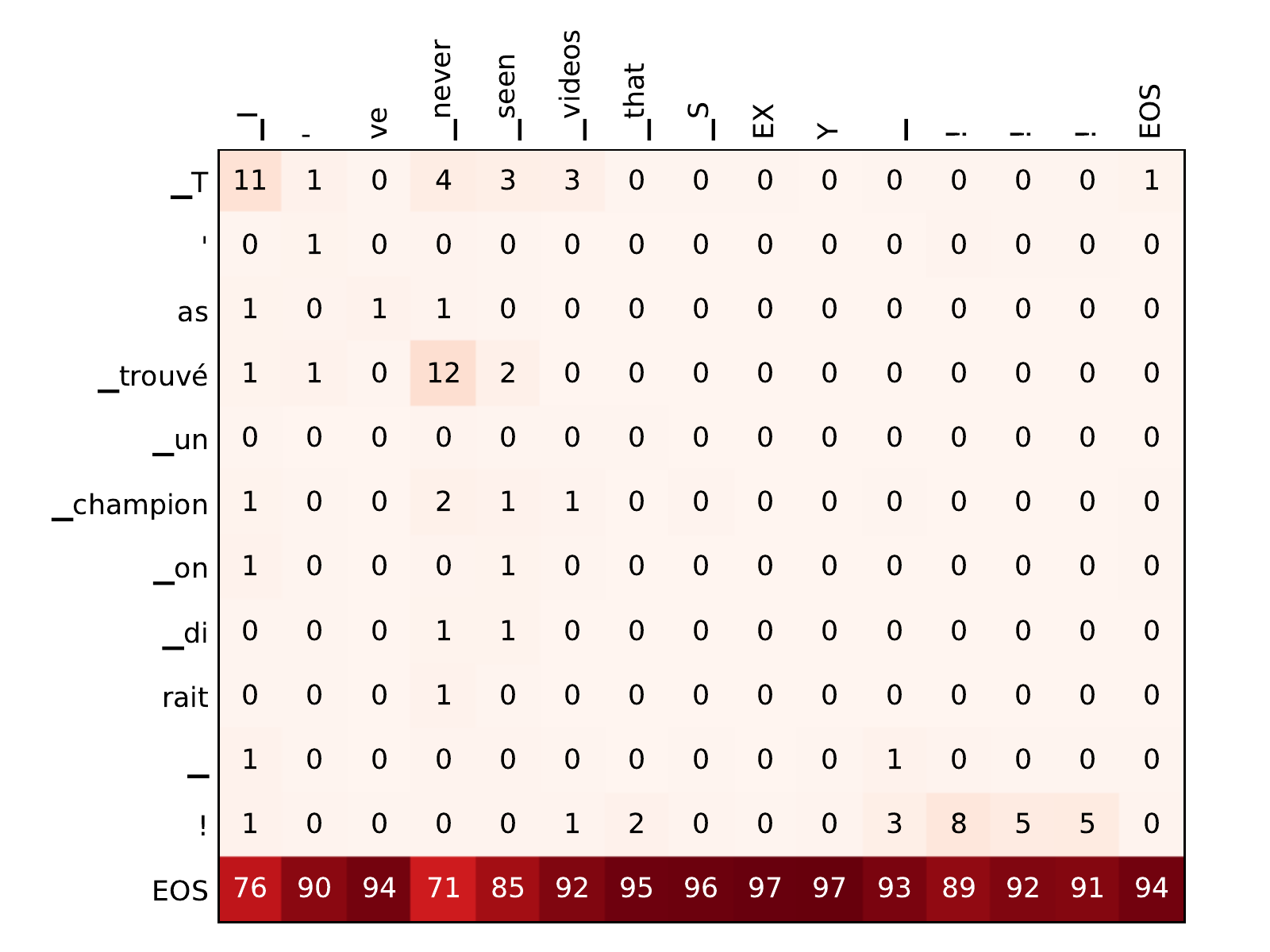}
	\caption{Attention matrix of a French (left) $\to$ English (top) Transformer when hallucinating. This is the average of the attention heads of the last decoder layer over the last encoder layer.}
	\label{figure:hallucination}
\end{figure}

We tried several approaches to eliminate hallucinations, whose corresponding scores are shown in Table~\ref{table:corpus_filtering}:
\begin{enumerate}[\hspace{0pt}1.]
	\setlength\itemsep{0em}	

\item Length filtering (removing any sentence pair whose length ratio is greater than 1.8, or 1.5 for CommonCrawl): removes most hallucinations and gives the best BLEU score (when combined with LID filtering). This type of filtering is common in MT pipelines \cite{koehn_2007}.

\item Excluding CommonCrawl from the training data: removes all hallucinations, but gives worse BLEU scores, suggesting that, albeit noisy, CommonCrawl is useful to this task.\footnote{And yet, CommonCrawl represents only 7.9\% of all lines and 6.5\% of all words in WMT.}

\item Attention-based filtering: we observed that when hallucinating, an NMT model produces a peculiar attention matrix (see Figure~\ref{figure:hallucination}), where almost all the probability mass is concentrated on the source \texttt{EOS} token. A similar matrix is produced during the forward pass of training when facing a misaligned sentence pair.
We filtered CommonCrawl as follows: we trained a baseline FR$\to$EN model on WMT without filtering, then translated CommonCrawl while forcing the MT output to be the actual reference, and extracted the corresponding attention matrices. We computed statistics on these attention matrices: their entropy and proportion of French words with a total attention mass lower than 0.2, 0.3, 0.4 and 0.5. Then, we manually looked for thresholds to filter out most of the misalignments, while removing as little correctly aligned data as possible.

\end{enumerate}

A combination of LID, length-based and attention-based filtering removed all hallucinations in the MT outputs, while obtaining excellent BLEU scores. The resulting corpus has 12\% fewer lines.\footnote{LID: -5\%, length filtering: -6.7\%, attention filtering: -0.5\%.} We use this filtered data for both FR$\to$EN and EN$\to$FR. As the JA$\leftrightarrow$EN training data seemed much cleaner, we only did a LID filtering step.

\begin{table}
\begin{tabular}{cccc|ccc}
	LID & Len & CC & Att & FR & Hallu. & BLEU \\
	\hline
	& & \checkmark & & 126 & 46 & 34.4 \\ 
	& & & & 0 & 12 & 34.8 \\ 
	\checkmark & \checkmark & & & 0 & 0 & 35.2 \\ 
	\checkmark & & \checkmark & & 0 & 29 & 37.7 \\ 
	\checkmark & \checkmark & \checkmark & \checkmark & 0 & 0 & 38.7 \\ 
	\checkmark & \checkmark & \checkmark & & 0 & 10 & 39.6 \\ 
\end{tabular}
\caption{Number of hallucinations and French-language outputs (according to \texttt{langid.py})  when translating MTNT-test (FR$\to$EN). LID: language identifier, Len: length filtering, CC: training data includes CommonCrawl, Att: attention-based filtering.}
\label{table:corpus_filtering}
\end{table}

\begin{table}
\begin{tabular}{l|p{6cm}}
	SRC & T'as trouv\'e un champion on dirait ! \\
	\hline
	REF & You got yourself a champion it seems ! \\
	MT & I've never seen videos that SEXY !!! \\
\end{tabular}
\caption{Example of hallucination by a FR$\to$EN Transformer trained on WMT15 data without filtering.}
\label{table:hallucination}
\end{table}

\paragraph{SentencePiece}

We use SentencePiece \cite{kudo_sentencepiece_2018} for segmentation into subword units.

An advantage of SentencePiece is that it does not require a prior tokenization step (it does its own coarse tokenization, based on whitespaces and changes of unicode categories). It also escapes all whitespaces (by replacing them with a meta symbol), so that its tokenization is fully reversible.
This is convenient for emoticons, which Moses-style tokenization tends to break apart irreversibly.

SentencePiece also normalizes unicode characters using the NFKC rules (e.g., $\frac{1}{2}\to$ 1/2).
It is useful for Japanese, which sometimes uses double-width variants of the ASCII punctuation symbols (e.g., ``fullwidth question mark'' in unicode table).

We tried different settings of SentencePiece, and settled with the BPE algorithm \cite{sennrich_neural_2016},\footnote{SentencePiece also implements ULM \cite{kudo_subword_2018}.} with a joined model of 32k tokens for FR$\leftrightarrow$EN (with a vocabulary threshold of 100), and two separate models of size 16k for JA$\leftrightarrow$EN.

\paragraph{Japanese tokenization}

SentencePiece's tokenization is based mostly on whitespaces, which are very rare in Japanese. For this reason, a pre-tokenization step may be useful (as a way to enforce some linguistic bias and consistency in the BPE segmentation).

We tested several tokenizers for Japanese: MeCab (with IPA and Juman dictionaries),\footnote{\url{http://taku910.github.io/mecab/}} Juman++,\footnote{\url{https://github.com/ku-nlp/jumanpp}} and Kytea.\footnote{\url{http://www.phontron.com/kytea/}} MeCab and KyTea gave comparable results, slightly better than when using no pre-tokenization (especially when Japanese is the target language), and Juman++ gave worse results. We settled with Kytea, which is the official tokenizer used on the EN$\to$JA task.\footnote{We use the default model shipped with KyTea.}

\subsection{Model and hyper-parameters}

We use Transformer Big for FR$\leftrightarrow$EN and JA$\to$EN, and Transformer Base for EN$\to$JA. We work with Fairseq, with essentially the same hyper-parameters as \citet{ott_2018}.

For FR$\leftrightarrow$EN, we fit up to 3500 tokens in each batch, while training on 8 GPUs (with synchronous SGD). We accumulate gradients over 10 batches before updating the weights. This gives a theoretical maximum batch size of 280k tokens. These models are trained for 15 epochs, with a checkpoint every 2500 updates. We set the dropout rate to 0.1. The source and target embedding matrices are shared and tied with the last layer.

For JA$\leftrightarrow$EN, we fit 4000 tokens in each batch, and train on 8 GPUs without delayed updates, for 100 epochs with one checkpoint every epoch. We set the dropout rate to 0.3.

For both language pairs, we train with Adam \cite{kingma_2015}, with a max learning rate of 0.0005, and the same learning rate schedule as \citet{ott_2018,vaswani_2017}. We also do label smoothing with a 0.1 weight. We average the 5 best checkpoints of each model according to their perplexity on the validation set. We do half precision training, resulting in a $3\times$ speedup on V100 GPUs \cite{ott_2018}.

\section{Robustness techniques}
\label{section:robustness_techniques}

We now describe the techniques that we applied to our baseline models to make them more robust to the noise found in MTNT.

\subsection{Case handling}

One of the sources of noise in the MTNT data is capital letters. On the Web, capital letters are often used for emphasis (to stress one particular word, or for ``shouting''). However, NMT models treat uppercase words or subwords as completely different entities than their lowercase counterparts. BPE even tends to over-segment capitalized words that were not seen in its training data.

One solution, used by \citet{levin_2017} is to do factored machine translation \citep{sennrich_linguistic_2016,garcia_martinez_2016}, where words (or subwords) are set to lowercase and their case is considered as an additional feature.

In this work, we use a simpler technique that we call ``inline casing'', which consists in using special tokens to denote uppercase (\texttt{<U>}) or title case subwords (\texttt{<T>}), and including these tokens within the sequence right after the corresponding (lowercase) subword. For instance, \texttt{They were SO TASTY!!} $\to$ \texttt{they \textless{}T\textgreater{} \_were \_so \textless{}U\textgreater{} \_tas \textless{}U\textgreater{} ty \textless{}U\textgreater{} !!}. SentencePiece is trained and applied on lowercase text and the case tokens are added after the BPE segmentation. We also force SentencePiece to split mixed-case words (e.g., \texttt{MacDonalds} $\to$ \texttt{\_mac \textless{}T\textgreater{} donalds \textless{}T\textgreater{}}).

\subsection{Placeholders}

MTNT contains emojis, which our baseline MT models cannot handle (unicode defines over 3~000 unique emojis).
We simply replace all emojis in the training and test data with a special \texttt{<emoji>} token.
Models trained with this data are able to recopy \texttt{<emoji>} placeholders at the correct position.\footnote{We ensure that there is always the same number of placeholders on both sides of the training corpus.} At test time, we replace target-side placeholders with source-side emojis in the same order.

We use the same solution to deal with Reddit user names (e.g., \texttt{/u/frenchperson}) and subreddit names (e.g., \texttt{/r/france}). MT models sometimes fail to recopy them (e.g., \texttt{/u/fran\c{c}ais}). For this reason, we identify such names with regular expressions (robust to small variations: without leading \texttt{/} or with extra spaces), and replace them with \texttt{<user>} and \texttt{<reddit>} placeholders.

\subsection{Natural noise}
We extract noisy variants of known words from the MTNT monolingual data, thanks to French and English lexicons and an extended edit distance (allowing letter swaps and letter repetitions). We also manually build a list of noise rules, with the most common spelling errors in English (e.g., your/you're, it/it's) and French (e.g., \c{c}a/sa, \`a/a), punctuation substitutions, letter swaps, spaces around punctuation and accent removal.
Then we randomly replace words with noisy variants and apply these noise rules on the source side of MTNT-train, CommonCrawl and News Commentary (MTNT-train, TED and KFTT for EN$\to$JA), and concatenate these noised versions to the clean training corpus.

\subsection{Back-translation}

Back-translation \citep{sennrich_improving_2016,edunov_2018} is a way to take advantage of large amounts of monolingual data. This is particularly useful for domain adaptation (when the parallel data is not in the right domain), or for low-resource MT (when parallel data is scarce).

In this task, we hope that back-translation can help on JA$\to$EN, where we have less parallel data, and on FR$\leftrightarrow$EN to expand vocabulary coverage (in particular w.r.t. recent named entities and news topics which are often evoked on Reddit).

Table~\ref{table:monolingual_data} describes the monolingual data which is available for constrained submissions. News-discuss (user comments on the Web about news articles) is probably more useful than news-crawl as it is closer to the domain.
We use our baseline models presented in Section~\ref{section:baseline_models} to back-translate the monolingual data. Following \citet{edunov_2018}, we do sampling instead of beam search, with a softmax temperature of $\frac{1}{0.9}$.

In all language pairs, we back-translate the target-language MTNT monolingual data, with one different sampling for each epoch. We also back-translate the following data:
\begin{itemize}
	\setlength\itemsep{0em}
	\item JA$\to$EN: $\frac{1}{20}$\textsuperscript{th} of \emph{news-discuss.en} per epoch (with rotation at the 21\textsuperscript{th} epoch).
	\item FR$\to$EN: $\frac{1}{5}$\textsuperscript{th} of \emph{news-discuss.en} per epoch (with rotation at the 6\textsuperscript{th} epoch).
	\item EN$\to$FR: \emph{news-discuss.fr} with one different sampling for each epoch and $\frac{1}{5}$\textsuperscript{th} of \emph{news-crawl.fr} (with rotation at the 6th epoch).
\end{itemize}

\subsection{Tags}
We insert a tag at the beginning of each source sentence, specifying its type: \texttt{<BT>} for back-translations, \texttt{<noise>} for natural noise, \texttt{<real>} for real data, and \texttt{<rev>} for MTNT data in the reverse direction (e.g., for JA$\to$EN MT, we concatenate MTNT JA$\to$EN and ``reversed'' MTNT EN$\to$JA). Like \citet{vaibhav_2019}, we found that ``isolating'' the back-translated data with a different source-side tag gave better BLEU scores. At test time, we always use the \texttt{<real>} tag.

Like \citet{kobus_2017}, we also use tags for domain adaptation. We prepend a tag to all source sentences specifying their corpus. For instance, sentences from MTNT get the \texttt{<MTNT>} tag and those from Europarl get the \texttt{<europarl>} tag. These ``corpus'' tags are used in conjunction with the ``type'' tags (e.g., MTNT back-translated sentences begin with \texttt{<MTNT> <BT>}). At test time, we use \texttt{<MTNT>} to translate MTNT-domain text, and no corpus tag to translate out-of-domain text.

We found that this method is roughly as good for domain adaptation as fine-tuning. We settle with corpus tags (rather than fine-tuning), as it is more flexible, less tricky to configure and has better properties on out-of-domain text.

\section{Results}
\label{section:results}

\begin{table}[t]
	\hspace{-.2cm}
	\begin{tabular}{lccc}
		Model & Test & Valid & Blind \\
		\hline
		MTNT & 6.7$^\dagger$ & -- & 5.8 \\
		MTNT fine-tuned & 9.8$^\dagger$ & -- & -- \\
		\hline
		Transformer base + tags & 13.5 & 11.2 & 13.7 \\ 
		+ Back-Translation (BT) & 15.0 & 12.8 & 14.1 \\ 
		+ Trans. big architecture $^{**}$ & 15.5 & 12.4 & 14.0 \\ 
		+ Ensemble of 4 $^*$ & \textbf{16.6} & \textbf{13.7} & \textbf{15.5} 
	\end{tabular}
	\vspace{-.05cm}
	\caption{BLEU scores of the JA$\to$EN models on MTNT-test, MTNT-valid and MTNT-blind.}
	\label{table:ja_en_bleu}
\end{table}

\begin{table}[t]
	\hspace{-.1cm}
	\begin{tabular}{lccc}
		Model & Test & Valid & Blind \\
		\hline
		MTNT & 9.0$^\dagger$ & -- & 8.4 \\
		MTNT fine-tuned & 12.5$^\dagger$ & -- & -- \\
		\hline		
		Transformer base + tags & 19.5 & 19.0 & 16.6 \\ 
		+ BT + natural noise $^{**}$ & 19.4 & 19.4 & 16.8 \\ 
		+ Ensemble of 6 $^{*}$ & \textbf{20.7} & \textbf{21.2} & \textbf{17.9} \\ 
	\end{tabular}
	\vspace{-.05cm}
	\caption{BLEU scores of the EN$\to$JA models.}
	\label{table:en_ja_bleu}	
\end{table}

\begin{table}[t]
	\centering
	\begin{tabular}{lcHcc}
		Model & Test & Valid & News & Blind \\
		\hline
		MTNT & 23.3$^\dagger$ & -- & -- & 25.6 \\
		MTNT fine-tuned & 30.3$^\dagger$ & -- & -- & -- \\
		\hline
		Transformer big & 39.1 & 27.3 & 39.3 & 40.9 \\ 
		+ MTNT + tags & 43.1 & 35.3 & 39.2 & 45.0 \\ 
		+ BT + natural noise $^{**}$ & 44.3 & 36.3 & 40.2 & 47.0 \\ 
		+ Ensemble of 4 $^{*}$ & \textbf{45.7} & \textbf{37.7} & \textbf{40.9} & \textbf{47.9} \\ 
	\end{tabular}
	\vspace{-.05cm}
	\caption{BLEU scores of the FR$\to$EN models on MTNT-test, news-test 2014 and MTNT-blind.}
	\label{table:fr_en_bleu}	
\end{table}

\begin{table}[t]
	\centering
	\begin{tabular}{lcHcc}
		Model & Test & Valid & News & Blind \\
		\hline
		MTNT & 21.8$^\dagger$ & -- & -- & 22.1 \\
		MTNT fine-tuned & 29.7$^\dagger$ & -- & -- & -- \\
		\hline		
		Transformer big & 33.1 & 31.3 & 40.7 & 37.0 \\ 
		+ MTNT + tags & 38.8 & 38.3 & 40.2 & 39.0 \\ 
		+ BT + natural noise $^{**}$ & 40.5 & 39.2 & 42.3 & 41.0 \\ 
		+ Ensemble of 4 $^{*}$ & \textbf{41.0} & \textbf{40.3} & \textbf{42.9} & \textbf{41.4} \\ 
	\end{tabular}
	\vspace{-.05cm}
	\caption{BLEU scores of the EN$\to$FR models.}	
	\label{table:en_fr_bleu}	
\end{table}

\vspace{-.1cm}
Tables~\ref{table:ja_en_bleu},~\ref{table:en_ja_bleu},~\ref{table:fr_en_bleu} and~\ref{table:en_fr_bleu} give the BLEU scores of our models on the MTNT-valid, MTNT-test and MTNT-blind sets (i.e., final results of the task). For FR$\leftrightarrow$EN we also give BLEU scores on news-test 2014, to compare with the literature, and to measure general-domain translation quality after domain adaptation. For news-test, we use Moses' \texttt{normalize-punctuation.perl} on the MT outputs before evaluation.

``MTNT'' and ``MTNT fine-tuned'' are the baseline models of the task organizers \cite{michel_2018}. The models marked $^{*}$ and $^{**}$ were submitted respectively to the competition as primary and secondary systems. Our primary ensemble models ranked first in all translation directions (with +0.7 up to +3.1 BLEU compared to the next best result). $\dagger$ means that different SacreBLEU parameters were used (namely ``intl'' tokenization).

The ``robustness'' techniques like inline casing, emoji/Reddit placeholders and natural noise had little to no impact on BLEU scores. They solve problems that are too rare to be accurately measured by BLEU.
For instance, we counted 5 emojis and 36 ``exceptionally'' capitalized words in MTNT-test. Improvements could be measured with BLEU on test sets where these phenomena have been artificially increased: e.g., an all-uppercase test set, or the natural noise of \citet{karpukhin_2019}.

Most of the BLEU gains were obtained thanks to careful data filtering and pre-processing, and thanks to domain adaptation: back-translation and integration of in-domain data with corpus tags.

\vspace{-.1cm}
\paragraph{Punctuation fixes}
We looked at the translation samples on the submission website, and observed that the French references used apostrophes and angle quotes. This is inconsistent with the training data (including MTNT), which contains mostly ASCII single quotes and double quotes. A simple post-processing step to replace quotes led to a BLEU increase of 5 points for EN$\to$FR.\footnote{The organizers and participants were informed of this.}

\vspace{-.1cm}
\section{Conclusion}

\vspace{-.1cm}
We presented our submissions to the WMT Robustness Task. The goal of this task was to build Machine Translation systems that are robust to the types of noise found on social media, in two language pairs (French-English and Japanese-English). Thanks to careful pre-processing and data filtering, and to a combination of several domain adaptation and robustness techniques (special handling of capital letters and emojis, natural noise injection, corpus tags and back-translation), our systems ranked first in the BLEU evaluation in all translation directions.

\clearpage
\bibliography{main}
\bibliographystyle{acl_natbib}

\end{document}